\documentclass[runningheads]{llncs}
\usepackage{graphicx}
\usepackage{amsfonts}
\usepackage{mathtools}
\usepackage{multirow}
\usepackage{tabularx}
\usepackage[ruled,linesnumbered]{algorithm2e}

\begin{document}
%

\title{STEP-EZ: Syntax Tree guided semantic ExPlanation for Explainable Zero-shot modeling of clinical depression symptoms from text}

%
%
\author{Nawshad Farruque\inst{1,2} \and
Randy Goebel\inst{1,2} \and
Osmar Za\"{i}ane\inst{1,2} \and
Sudhakar Sivapalan\inst{3}
}
\authorrunning{Farruque, N et al.}
%
\institute{Department of Computing Science, Faculty of Science, University of Alberta \and
Alberta Machine Intelligence Institute (AMII), University of Alberta
\email{\{nawshad,rgoebel,zaiane\}@ualberta.ca}\\ \and
Department of Psychiatry, Faculty of Medicine and Dentistry, University of Alberta \\
\email{sivapala@ualberta.ca}}
\maketitle              
\begin{abstract}
We focus on exploring various approaches of \emph{Zero-Shot Learning} (ZSL) and their explainability for a challenging yet important supervised learning task notorious for training data scarcity, i.e. \emph{Depression Symptoms Detection} (DSD) from text. We start with a comprehensive synthesis of different components of our ZSL modeling and analysis of our ground truth samples and Depression symptom clues curation process with the help of a practicing clinician. We next analyze the accuracy of various state-of-the-art ZSL models and their potential enhancements for our task. Further, we sketch a framework for the use of ZSL for hierarchical text-based explanation mechanism, which we call, \emph{Syntax Tree-Guided Semantic Explanation} (STEP). Finally, we summarize experiments from which we conclude that we can use ZSL models and achieve reasonable accuracy and explainability, measured by a proposed Explainability Index (EI). This work is, to our knowledge, the first work to exhaustively explore the efficacy of ZSL models for DSD task, both in terms of accuracy and explainability.

\end{abstract}
\section{Introduction}
\label{sec:intro}
Depression has been found to be the most common psychiatric disorder amongst young adults, accounting for 75\% of all clinical health care admissions in developed countries \cite{boyd1982screening}. Unfortunately, due to stigma around mental health disorders, people with Depression often hide their condition or ignore it, which further aggravates their condition and results in deeper societal problems, including substance abuse, self harm and even incidences of suicide. Earlier studies have shown that, young people who are suffering from Depression, often show help-seeking behavior by speaking out through social media posts. There have been ample research  done to date \cite{Choudhury2013Pred,coppersmith2015clpsych,moreno2011feeling,de2014mental,Rude2004} that successfully lays the foundation of using social media posts as a proxy for a social media user's daily language usage, then analyzing posts to extract signs of Depression with reasonable accuracy. However, to detect and confirm signs of Depression in the most clinically accurate way, a clinician needs to uncover clinical symptoms exhibited by an user on a day-to-day basis. 

There has been research, such as \cite{mowery2017understanding} and \cite{cheng2016psychologist}, on identifying signs of clinical Depression symptoms in social media posts, such as Tweets or Facebook posts  as a first and important step to establish user-level Depression detection. However, the most challenging aspect of these research is the paucity of valid data, i.e., creation and curation of data-sets validated by clinical Psychiatrists, who see patients and analyze their language in their daily practice. Since collecting large volumes of human annotated data is a daunting task, we exploit powerful language models available these days: here we concentrate on using these pre-trained language and natural-language-inference task models (including their learned representations and few enhancements over those representations) to formulate our Zero-Shot Learning (ZSL) approach for detecting clinical clues of depressive symptoms from Tweets. Zero-shot learning is a machine learning setup allowing between-class attribute transfer in order, at test time, to observe and predict the classes of samples from classes that were not observed during training. This is generally achieved by associating observed and non observed classes through the encoding of observable distinguishing properties of objects \cite{larochelle2008zero}. 
To dig deep in the performance of these models, accuracy alone is not sufficient, therefore we propose a constituency/syntax tree-guided semantic explanation mechanism, which ensures multiple, minimal and meaningful explanations over fixed length texts and less meaningful explanations. Further we propose an \emph{explanation index} (EI) to score the quality of our explanations and analyze all our ZSL models based on their explainability efficacy reflected through our EI-Score. In summary our contributions are as follows:
\begin{enumerate}
    \item We use state-of-the-art language models, their learned representations and a few subject matter techniques to augment those representations to build our ZSL framework.
    \item Since a ZSL task requires minimal clues that can help it to label an ``unseen'' sample, we carefully curate the clues of depressive symptoms with the help of a practicing clinician,  Diagnostic and Statistical Manual (DSM-5) for Depression and ``clinician-reported'' Depression questionnaire. This means that the identified clues directly reflect the signs of Depression symptoms in an individual's language usage from the clinician's point of view. 
    \item We propose a text explainability algorithm called \emph{Syntax Tree-Guided Semantic Explanation} (STEP), that encourages multiple short and hierarchical phrases inside a Tweet to explain its label.
    \item In companion with the previous point, we propose an \emph{Explainability Index} (EI) that is used to grade the explainability mechanism.
    
According to our knowledge, this is the first work of its kind that comprehensively analyzes the capability of ZSL models, from detecting signs of clinical Depression symptoms from text. It is also possible to extend this work to complement active learning frameworks, and thus gather more data for this kind of research.
\end{enumerate}

\section{ZSL Model Preliminaries}
\label{section:zsl-concepts}
Given, a Tweet, $T$, 
it has a label, $L_{i}$ where, $L_{i} \in \{L_{1}, ... L_{m}\}$ if it has a strong \textit{membership-score} with any of its descriptors, $l_{j}$ where, $l_{j} \in L_{i}$ and $L_{i}$ = $\{l_{1}, ... l_{n}\}$.  Here the descriptor $l_{j}$ is a representation of the label $L_{i}$. For example, consider that one of our Depression symptoms $L_{i}$ is  ``Low Mood'' and the descriptors representing $L_{i}$ is a set, $l$ = $\{Despondency, Gloom, Despair\}$. If $T$ has a strong membership-score with any members of $l$, we can say $T$ has the label $L_{i}$ = ``Low Mood.'' Since our DSD task is a multi-label classification task, it is possible for $T$ to have multiple labels at the same time, because it can have a strong membership-score with respect to the descriptor(s) of multiple labels. This is a ZSL paradigm, because we determine the label(s) of $T$ based on its membership-score with respect to any of the descriptors in $l$ at test time, with which our models are probably not familiar with at training time. 

We use mainly two broad families of ZSL models in this paper, such as, embedding (both sentence and word) models and Natural Language Inference (NLI) pre-trained models. For embedding models, we represent $T$ and each of the $l_{j}$'s using various classic and state-of-the-art word and sentence embedding models and measure their membership-scores based on how close they are in the vector space or cosine distance. For NLI models, we extract the probability type of entailment-scores which shows the membership-score for a $T$ with respect to each of the $l_{j}$'s. 

\begin{figure}[!ht]
\centering
\includegraphics[width=0.75 \textwidth]{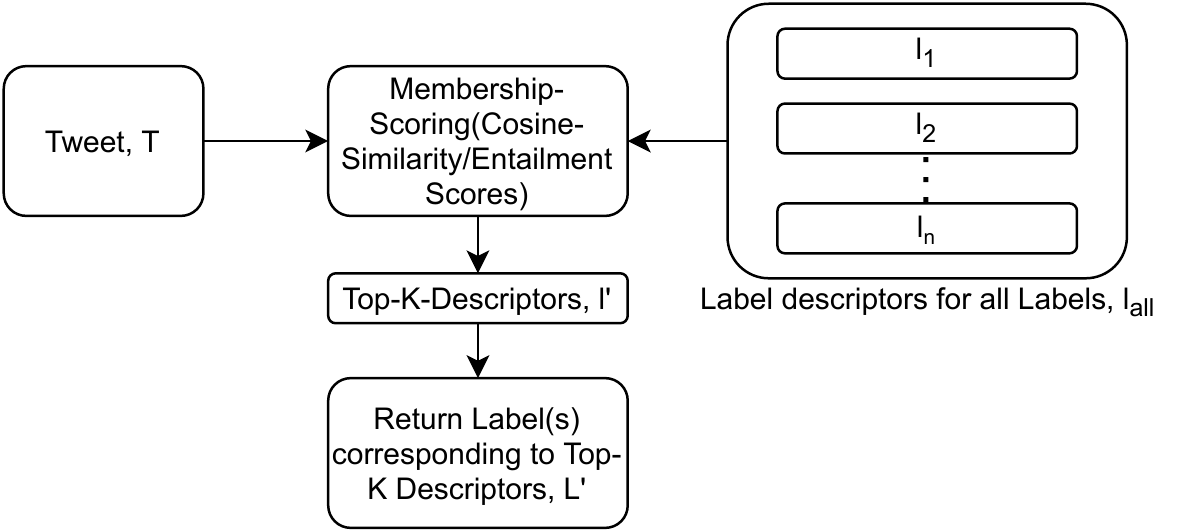}
\caption{\label{fig:ZSL-Framework}An overview of ZSL Framework}
\end{figure}

So to describe our ZSL framework, see Figure \ref{fig:ZSL-Framework}, we start with discussing our Ground Truth data curation process which is used to evaluate the accuracy of our models, see Section \ref{subsec:gt-curation}, next we discuss, Depression symptoms label $L$ and their descriptors ($l$) curation process, see Section \ref{subsec:label-curat}, later, we describe representation building of our Tweets and labels for Embedding families of models, see Section \ref{subsec:representation}, finally we discuss membership-score calculation between Tweets and Labels, see Section \ref{subsec:label-membership}.  We describe each of these processes, using the notation described above. Please note, in the following sections and through-out the paper, we use the term ``Embedding'' to define a set of word vectors. 

\subsection{Ground Truth Dataset (GTD) Curation}
\label{subsec:gt-curation}
Our \emph{Ground Truth Dataset} (GTD) 
for Depression symptoms was collected by a research group at the University of Montpellier \cite{vioules2018detection}.  They first curated a list of keywords which express signs of Depression, Self-harm and Suicidality. Later they confirmed alignment with Twitter users whose profile description matches 
with those keywords and who frequently post Tweets containing those keywords. Finally, with the help of three annotators, including psychologists, they curated 500 Tweets with depressive and non-depressive labels.
We use GTD and label its entries for nine clinical Depression symptoms as per the Diagnostic and Statistical Manual version 5 (DSM-5) \cite{DSM-5}. Before annotating the dataset, we first identify confusing Tweets which could be member of several Depression symptom categories. Later, we arrange a one-on-one discussion session with a clinician, to confirm possible membership of those Tweets for one or more Depression symptoms categories according to several Depression questionnaires, such as, PHQ-9, MADRS, BDI and CES-D \cite{national2010classification}. Finally, based on the clinical insights combined with questionnaire concepts, we then annotate 256 Tweets for Depression symptoms. We annotate the same dataset twice by the same annotator and achieve a 0.85 test-retest \cite{guttman1945basis} reliability coefficient score.

\subsection{Labels(L) Curation for ZSL}
\label{subsec:label-curat}
First we use well known Depression questionnaires and align the common Depression symptoms concepts with the help of the clinician, to better understand the general language used for describing individual clinical symptoms. We then separate the minimal description of the symptoms or \textbf{Header} for DSM-5, \textbf{DSM-Header (DH)} and MADRS \cite{hollandare2010comparison},  \textbf{MADRS-Header(MH)} and slightly elaborated description of the symptoms, or \textbf{Lead} for MADRS, \textbf{MADRS-Lead(ML)}. We curate a list of elaborated descriptions of Depression symptoms concept with the help of all the questionnaires (as listed in the previous section), and discussion with the clinician, which we call \textbf{All}. In addition, we combine only the headers of DSM-5 and MADRS for corresponding Depression symptoms, which we call \textbf{MADRS+DSM-Header(MH+DH)}. Finally, we use a hand curated and expert annotated Depression symptoms lexicon, named \textbf{SSToT} \cite{yazdavar2017semi}. Our labels are mostly based on DSM-5 and MADRS, because DSM-5 is the manual used for clinical Depression detection world-wide, and the basis for most of the other developed Depression questionnaires,  and MADRS, because it is  administered to clinicians for clinical Depression detection. Since we use annotation advice from the clinician to annotate our data and we want to analyze the Tweets for Depression symptoms from the clinician's point of view, MADRS perfectly fits our need; its Headers and Leads are more easily understandable by the annotators compared to other questionnaires. See Table \ref{tab:headers} for a sample for headers and leads for couple of Depression symptoms.

\begin{table}[h!]
\small
\centering
\caption{\label{tab:headers}A glimpse of few Depression symptoms labels ($L)$ and some Headers and Leads that constitute our $l_{all}$}
 \begin{tabular}{||p{3cm} | p{3cm} | p{3cm} | p{3cm} ||} 
 \hline
 Sample of Depression Symptoms & DSM-Headers (DH) & MADRS-Headers (MH) & MADRS-Leads (ML) \\ [0.5ex] 
 \hline\hline
 Disturbed sleep & Insomnia, Hypersomnia & Reduced sleep & Reduced duration of sleep, Reduced depth of sleep\\
 \hline
 Anhedonia & Loss of interest, Loss of pleasure & Inability to feel & Reduced interest in surroundings, Reduced ability to react with adequate emotion\\ [1ex] 
 \hline
 \end{tabular}
\end{table}

\subsection{Representation of Tweets and Labels for ZSL}
Here we separately discuss about various embedding based representation techniques for our Tweet, $T$ and Depression symptoms label descriptor, $l_{j}$.
\label{subsec:representation}
\subsubsection{Word-Embedding-Family(WEF)}\label{subsection:wef}
We use several classic word embedding models, including Google News (Google) \footnote{https://code.google.com/archive/p/word2vec/}, Twitter Glove (Glove) \footnote{https://nlp.stanford.edu/projects/glove/}, Twitter Skip-gram Embedding (TE) \cite{godin2015multimedia}, Depression Specific Embedding (DSE) trained on Depression specific corpora \cite{farruque2019augmenting}, Depression Embedding Augmented Twitter Embedding (ATE) \cite{farruque2019augmenting}, NLI pre-trained 
Roberta Embedding (Roberta-NLI) \cite{liu2019roberta} and Universal Sentence Encoder Embedding (USE) \cite{cer2018universal}. All these embeddings except DSE have been trained on millions of tokens. As USE and Roberta-NLI are sentence embeddings, we take the sentence vector for each word as their word vector. In the following sections, we describe how we leveraged them to build sentence representations. 

\subsubsection{Average Word Vector Models(WV-AVG)}
If we assume a Tweet, $T$ or a label descriptor, $l_{j}$ (see Section \ref{section:zsl-concepts}) as a our sentence, and each sentence, $S$ consists of $n$  words, i.e., $S = \{W_{1}, ... W_{n}\}$, \textit{``wv''} is a function that returns the vector representation of a word, then a sentence as an averaged word vector can be expressed as follows: 

\begin{equation}
\label{eq:wv-avg}
    \frac{\sum_{i=0}^{n}wv(W_{i})}{n}
\end{equation}

\subsubsection{Word Vector Mapper Models(WV-MAPPER)}
As originally proposed in \cite{farruque2019augmenting}, we learn a least square projection matrix, $M_{w}$, between the word vectors of the common vocabulary $V$ of both source and target embeddings. This learned matrix is then used to adjust word vectors of source embedding,  then later used to build WV-AVG sentence representation as outlined in Equation \ref{eq:wv-avg}. This method has been previously found to be effective for the depressive post detection task \cite{farruque2019augmenting}. For our WV-MAPPER models, our source embedding is one of the general Twitter pre-trained embedding (TE) and the target is the Depression Specific Embedding (DSE). The specification for this mapping is as follows, where $M_{w}^\ast $ is the learned projection matrix and $V_{S}$ and $V_{T}$ are the common vocabularies of source and target embeddings.

\begin{equation}
\label{eq:we-mapper}
    M_{w}^\ast = \arg\min || wv(V_{S})^\top M_{w} - wv(V_{T}) ||^2 
\end{equation}

\subsubsection{Sentence Embedding Family (SEF)}
\label{subsection:sent-embed}
We use state-of-the-art Roberta-NLI and USE sentence embeddings which are transformer based models and multi-task pre-trained on NLI and semantic textual similarity tasks (STS) (i.e. Roberta-NLI) and sentiment analysis tasks as well (i.e USE).

\subsubsection{Vanilla Sentence Vector Models(SV)}
Provided a sentence, S, its sentence vector is represented as $sv(S)$.
\subsubsection{Sentence-to-Word Vector Mapper Models(SV-WV-MAPPER)}\label{subsubsec:sv-wv-mapper}
We use the same formulation as stated in Equation \ref{eq:we-mapper}, however, while learning the projection matrix $M_{s}^\ast$, here we use the sentence vector of a source word and learn its projection to word vector of the target word for the common vocabulary between the source and target embeddings.  All the other notations are the same as noted earlier:
\begin{equation}
\label{eq:se-mapper}
    M_{s}^\ast = \arg\min || sv(V_{S})^\top M_{s} - wv(V_{T}) ||^2 
\end{equation}

Later, we use $M_{s}^\ast$ to transform $sv(S)$. The intuition behind this mapping is that a sentence vector is not good at representing a label descriptor $l_{j}$, which are usually short-phrases. So we project a sentence vector for a Tweet, $T$ and its corresponding label descriptor, $l_{j}$ to a word-vector space (in our case ``DSE'', where we have salient semantic clusters of depressive symptoms). 

\subsection{Natural-Language-Inference (NLI) Model}
We use the Facebook-BART \cite{lewis2019bart} model, which uses BERT and GPT hybrid pre-training on NLI task and performs very well in ZSL settings. In the NLI task, a model is given a premise and a hypothesis and it needs to predict whether the hypothesis entails the premise or contradicts it. It has been found to be a very effective ZSL model, which can pretty accurately predict whether a given label (in our case label descriptor, $l_{j}$) entails a particular sample, in our case a Tweet, ($T$). This mechanism provides a probability score $\in [0,1]$ of entailment for each label.

\subsection{ZSL Top-K-Label-Membership Formulation}\label{subsec:label-membership}

At the heart of our Top-K-Label-Membership formulation is an algorithm that determines the membership of a Tweet, $T$ with all the descriptors, $l_{all}$ for all labels, $L$. We later sort the descriptors based on their membership-scores with $T$ in descending order (assuming higher score means better membership), 
and get $l_{all-sorted}$ (see Algorithm \ref{algo:descriptor-score}). Finally, we return the labels, $L' \subset L$  represented by the top-k descriptors, $l' \subset l_{all-sorted}$ as our candidate labels for the Tweet, $T$  (see Algorithm \ref{algo:label-predictor}). In the following sections we describe our membership-scoring details for Embedding and NLI family of models.

\begin{algorithm}[H]
\small
    \KwData{$T$, $l_{all}$, $mode$}
    \KwResult{$l_{all-sorted}$}
    $l_{all-sorted} \leftarrow \emptyset$ \; 
    membership-score-dictionary $\leftarrow \emptyset$ \;
    \If {$mode$ is ``Embeddings''}{
        \ForEach{$l \in l_{all}$}{
            membership-score-dictionary$[l] \leftarrow$ 1 - cosine-distance$(T, l)$   \;
        }
    }
    \ElseIf{$mode$ is ``NLI''}{
        \ForEach{$l \in l_{all}$}{
            membership-score-dictionary$[l] \leftarrow$ entailment-prob-score$(T, l)$ \;
        }
    }
    $l_{all-sorted} \leftarrow$ descriptors(sort-desc(membership-score-dictionary)) \;
    return $l_{all-sorted}$ \;
    \caption{\label{algo:descriptor-score}Sorted-Descriptors}
\end{algorithm}

\begin{algorithm}[H]
\small
    \SetAlgoLined
    \KwData{$L$, $l_{all-sorted}$, $k$}
    \KwResult{$L'$}
     $L' \leftarrow \emptyset$ \;
     $n \leftarrow 0$ \;
     \While{$n < k$}{
         \ForEach{$l' \in l_{all-sorted}$}{
         $i \leftarrow 0$ \;
                \ForEach{$L_{i} \in L$}{
                    \If{$l' \in L_{i}$}{
                        $L' \leftarrow L' \cup L_{i}$
                    }
                $i \leftarrow i + 1$
                }
            $n \leftarrow n + 1$
            }
    }
    return $L'$ \;
    \caption{\label{algo:label-predictor}Label-Predictor}
\end{algorithm}

\subsubsection{Embedding Family Models}
\begin{figure}[!ht]
\centering
\includegraphics[width=0.75 \textwidth]{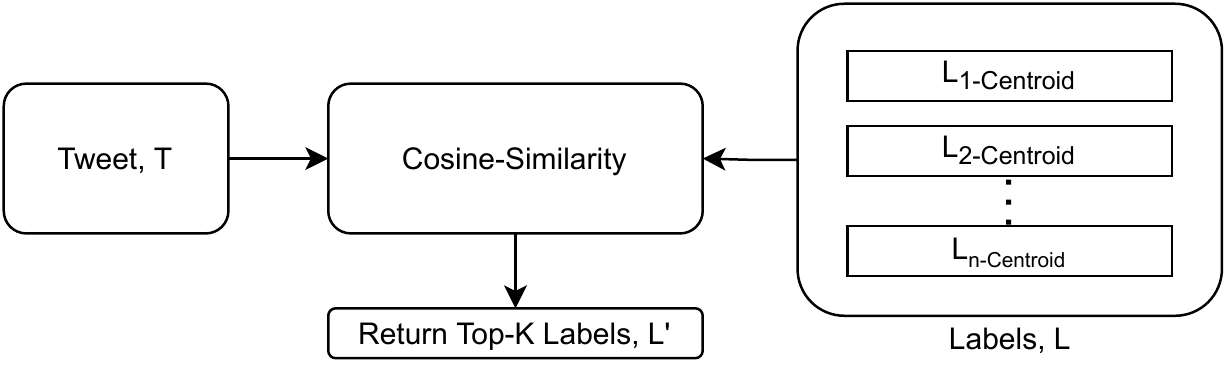}
\caption{\label{fig:ZSL-Centroids}An overview of ZSL-Centroid Methods}
\end{figure}

For this family of models, we use cosine similarity or $(1$ - cosine-distance$)$ to determine the membership of a vector representation of Tweet, $T$ to the same of any of the descriptors in $l_{all}$. 
\textbf{Centroid Membership:} In this scheme, we represent each label, $L_{i}$ with the average representation vectors of all of its descriptors, which we call ``centroid''. For example, in the centroid-based method, $T$ has label $L_{i}$ if $T$ has a strong membership-score with $centroid(L_{i})$, where $L_{i}=l=\{l_{1}, l_{2},..., l_{n}\}$ and $l$ is the set of descriptors. Then we return $L' \subset L$, i.e. the top-k labels, based on the descending order of the cosine similarity with $T$, as described earlier, as candidate labels for $T$. \textbf{Top-k Centroid Membership:} Similar to centroid membership, instead of considering all the descriptors of $L_{i}$, we use the Top-K descriptors based on the cosine similarity. See Figure \ref{fig:ZSL-Centroids} for an overview of centroid methods. \textbf{NLI Family Models:} As mentioned in Section \ref{subsection:sent-embed}, NLI models provide probability scores for entailment for a Tweet, $T$ to its descriptor, $l_{j}$. We follow a similar procedure as for the embedding family models except we use the entailment probability scores to find the final candidate labels, $L'$ for $T$.




\section{ZSL Model Explainabilty}
In this section we discuss our efforts to analyze the performance of ZSL models based on their explainability. The core idea behind our explainability algorithms is to find which phrases express the same Depression symptoms label as the Tweet. Moreover, we would also like to determine how those phrases are surrounded by their neighbouring phrases to contribute to the semantics of a label. To do so, we propose two algorithms as described in the following sections. Our first algorithm ``Syntax Tree-Guided Semantic Explanation (STEP)'' respects the syntax tree-based compositionality to explore n-grams inside the Tweet, This compositionality may also contribute to the Tweet having a particular label (see Section \ref{subsec:step}). Our second algorithm, we call  n-gram based explanation (ngramex), naively divides a Tweet in its constituent n-grams (where ``n'' is pre-determined) to help explain a Tweet for its label (see Section \ref{subsec:ngramex}). Finally, in  Section \ref{subsec:ei} we propose an Explanation Index (EI) function that provides higher scores for multiple minimal explanations for a Tweet-Label as opposed to single or lengthy explanations. 

\subsection{Syntax Tree Guided-Semantic Explanation (STEP)}
\label{subsec:step}
The intuition behind STEP is, according to generative linguistic theory by Chomsky \cite{osterhout2012neurobiology} to understand the meaning of a sentence, is that humans combine words in at least two levels, such as a syntactic level and a semantic level. Since explaining a sentence requires understanding its semantics through syntactic composition, we reflect this theory in our explanation algorithm. First, we start by approximating semantic understanding of a Tweet, $T$ as a whole (or the label/Depression symptoms expressed by it), then we gradually explore the nodes of the syntax tree for $T$ in breadth-first manner and find out which n-grams (children of those nodes) also express the same, until all the nodes have been traversed. Finally, STEP returns the set of n-grams (where ``n'' is dynamic and $n \in \mathbb{Z}^+$) or ``explanations,'' $E$, in descending order of membership-score with the Tweet label. Here, we use the Algorithms \ref{algo:descriptor-score} and \ref{algo:label-predictor} for finding out the candidate label of n-grams at each node of the syntax tree. It is to be noted, we consider the most expressed candidate label (i.e. the candidate label which has the highest membership-score with the label of $T$) for an n-gram, instead of multiple labels for the ease of understanding our explainability mechanism.

\begin{algorithm}[H]
\small
    \SetAlgoLined
    \KwData{$T$}
    \KwResult{$E$}
    $Tree \leftarrow$ Syntax-Tree(T) \;
    Tweet-Label $\leftarrow$ Label(T) \;
    Explanation-Dictionary $\leftarrow \emptyset$ \;
    \While{$not$ Tree.traversedAllNode()} { 
    \tcp*[l]{Traversing the $Tree$ in Breadth-First order and from left-to-right nodes}
        \ForEach{$node \in Tree$ }
        {
            node-Label $\leftarrow$ Label(n-gram(node)) \;
            \If {node-Label == Tweet-Label}{
                node-Score $\leftarrow$ Score(n-gram(node), Tweet-Label) \;
                Explanation-Dictionary[n-gram(node)] $\leftarrow$ node-Score
            }
        }
    }
    $E \leftarrow$ explanations(sort-descending(Explanation-Dictionary) \;
    return $E$
    \caption{\label{algo:step}STEP}
\end{algorithm}

It is easy to see that we could use this process for each of the candidate labels for explainability analysis if needed. The entire process is described in Algorithm \ref{algo:step}. Further for the sake of simplicity, let us assume a function ``n-gram'' takes the leaves of each node and returns the corresponding n-gram; function ``Label'' takes an n-gram and returns its most expressive label; function ``Score'' returns the membership-score of the n-gram and the label, and the function ``Syntax-Tree'' returns, as the name suggests, a syntax tree of the corresponding n-gram. 


\subsection{N-gram based Explanation(ngramex)}
\label{subsec:ngramex}
In this algorithm, we simply partition $T$ into some pre-defined length of n-grams. Later we identify n-grams which have the same label as $T$, and return the list of those n-grams according to the descending order membership-score with a label, $T$.  

\section{Experimental Design}
We design our experiments to enable analysis with respect to model accuracy and explainability. We report two experiments to confirm the accuracy of our models, such as in (1) \textbf{Depression Symptoms Detection from Tweets (DSD)} task, which is our core task and (2) \textbf{Depressive Post Detection (DPD)} task, which confirms the predictive capability of our models in general to identify depressive 
vs. non-depressive Tweets. 
In terms of explainability, we formulate an explanation index (EI) and analyze how different models perform in terms of EI. 

\subsection{Train and Test Data-sets}
For our DSD task, We have 256 Tweets annotated with expert insight for nine clinical Depression symptoms except the symptom ``Retardation'' because our SSToT lexicon does not contain any samples for that category so for fair comparison we did not consider it. Since we use a supervised classification baseline, we randomly split our data-set into 80\% train-set ($\approx 205$ Tweets) and 20\% ($\approx 50$ Tweets) test-set. We have three such partitions because the experiments are expensive, and we report our results based on average F1 accuracy measures and their std. deviations on these test-sets individually for our top-k$=\{1,3,6,9\}$. 

For DPD task, we use rigorously human annotated 500 Depressive and non-Depressive Twitter posts. We partition it to 30 stratified train-test splits, since the experiment is fairly fast. Later, we report our F1-score averaged (with std. deviations) on the performance on these 30 sets of test-sets.

\subsection{Accuracy Scores}
Since our DSD task is a multi-label classification task and our data is rather imbalanced, we use a standard Micro-F1 score to evaluate our ZSL models. For the binary depressive posts detection task, we use a standard F1-score. 

\subsubsection{Depressive Symptoms Detection (DSD) Task}
We perform experiments on all the combinations of our ZSL models, such as, Google, Glove, Twitter, DSE, USE, Roberta-NLI plain and mapper methods and Facebook-BART.  The combinations arise from various Depression symptoms clues and curation strategies, such as (DSM-Header, MADRS-Header, DSM+MADRS-Header, MADRS-Lead, SSToT and All. In addition, we run these experiments for various configurations of top-k descriptors. However, to analyze and discuss our results, we report the best models under each of the ZSL family and their sub-families averaged over our 3 randomized test sets, see Table \ref{tab:dsd-f1-score}. For the baseline, we use BERT (``bert-base-uncased'' under ``Hugging-Face'' transformer library \footnote{https://huggingface.co/transformers/})  fine-tuned for our Depression symptoms dataset. 

The naming convention we follow for our Table \ref{tab:dsd-f1-score} is as follows, we use the embedding/nli model name followed by questionnaire-description-mode name and finally number of the Top-k descriptors. For example, for DSE(MH+DH)-Top-1, DSE is the name of the embedding model, (MH+DH) means MADRS+DSM-Headers and Top-1 indicates top-1 descriptor is used to find the Tweet's label.


\subsubsection{Depressive Post Detection (DPD) Task}
For this task, we use our membership-scores for various symptoms as the feature representation for the Tweets,  then send that to an SVM classifier and compare its performance with LIWC \cite{Pennebaker2003} feature representation based SVM classifier along with random uniform and all-majority-class baselines. We use an SVM classifier because it is found to be best performer, given our small dataset. 


\subsection{Explanation Index Score (EI-Score)}
\label{subsec:ei}
We propose an Explanation Index (EI) score to evaluate our ZSL Models in terms of their explainability. We report EI scores for both STEP and ngramex, and analyze their agreement
over different samples to compare and contrast. Let us assume a set of explanations, $E = \{e_{1},e_{2},....e_{n}\}$ for a particular Tweet for its label. Each $e_{i}$ corresponds to an n-gram explanation of a Tweet for its label. A function ``length'' returns the number of words in $e_{i}$, and the function ``rank'' returns the rank of a particular $e_{i}$ in $E$. Since $e_{i}$'s are in sorted order under $E$, the lower the rank the better the explanation. We can express our EI-Score for $E$ as follows,
\begin{equation}
    \frac{\sum_{i=0}^{n}EI_{i}}{n}
\end{equation}
where,
\begin{equation}
    EI_{i} = LengthScore(e_{i}) * RankScore(e_{i}) * Relevance(e_{i})
\end{equation}
\begin{equation}
    LengthScore(e_{i}) = 1 - (length(e_{i})/length(Tweet))
\end{equation}
\begin{equation}
    RankScore(e_{i}) = 1 - (rank(e_{i})/n)
\end{equation}
\begin{equation}
    Relevance =
    \begin{cases*}
      1 & if $Label(Tweet) == Label(e_{i})$ \\
      0        & otherwise
    \end{cases*}
\end{equation}

We can see that EI scores are higher for multiple explanations over a single explanation, and short explanation over lengthy explanations. It is possible that ngramex with a certain ``n'' can have a better score according to this scoring system, however, ngramex has a high possibility of returning non-salient explanations which are not useful to humans (see Table \ref{tab:exp-examples} in Section \ref{subsubsec:ei-score}). The range of our EI-Score function is between 0 and 1 and the higher score indicates the better explanation. We report two kinds of analyses here, such as, (1) regarding average EI-score for both STEP and ngramex(with n=3, because we found it's often scores better) over one of our test-set for correctly predicted samples by a ZSL model. In Table \ref{tab:avg-ei-score} we report these average EI-Scores for our Top-3 ZSL models to analyze their explainability performance and (2) regarding the analysis of the rationale behind disagreement between STEP and ngramex EI-score, for that, we sort out few examples where STEP EI-Score is greater than the same for ngramex and vice-versa. Finally, we would like to mention that, EI-Score is not a fool proof scoring system. It scores higher for smaller explanations even though those may not make sense. Quality of explanation depends on both underlying ZSL model and the explanation mechanism. Here STEP works as a layer to reduce gibberish explanations as opposed to ngramex. 

\section{Results Discussion} 
Here we discuss the performance of our models based on two broad categories of performance measures, such as, (1) accuracy and (2) explainability \cite{atakishiyev2020multi} as follows,

\subsubsection{Depression Symptoms Detection (DSD) Task Accuracy} 



\begin{figure}[!ht]
\centering
\includegraphics[width=0.80 \textwidth]{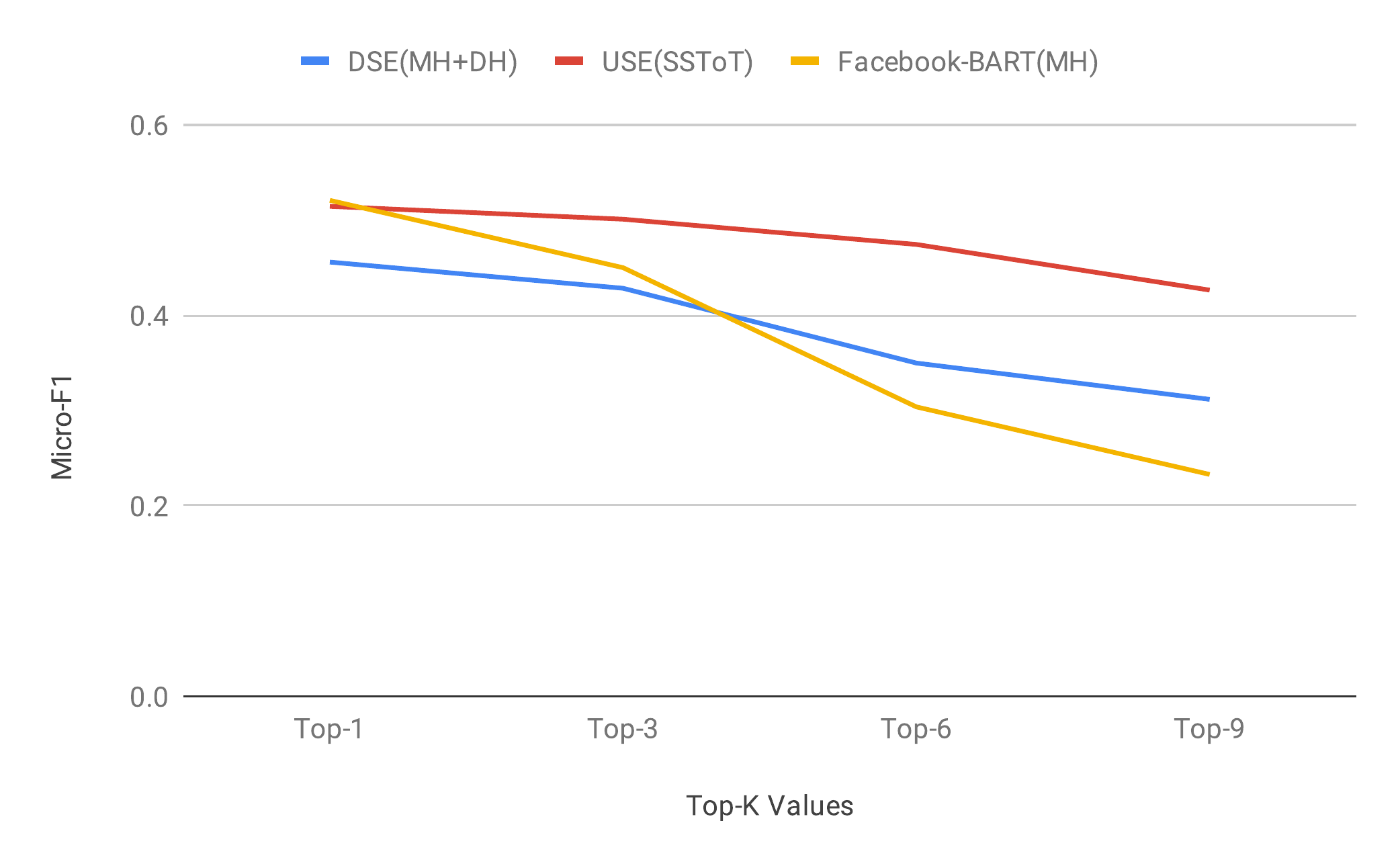}
\caption{\label{fig:top-k-effects}Effect of Top-K values on accuracy for the best performing models}
\end{figure}

We observe that NLI models are the best and Sentence Embedding Familty (SEF) models are on par indicates, NLI and sentence embedding models with their semantic similarity pre-training, are inherently better in ZSL tasks.  With respect to top-k we can see gradual decrease of Micro-F1 with the increase of top-k with pick at 1, see Figure \ref{fig:top-k-effects}. So for all the methods, first few most salient descriptors and their corresponding labels are more predictive. Better Micro-F1 also indicates that these classifiers are better at classifying majority class. Possibly majority classes are more straight-forward than their minority counter-parts and ZSL can also easily identify those classes. 

All the ZSL models have significantly better accuracy than supervised BERT-fine-tuned model which gives us a hope that we can start with ZSL to gather more data before training supervised models. 

Most of the ZSL do not require exhaustive list of labels, such as ``SSToT'' (except, USE) or ``All'' as described in Section \ref{subsec:label-curat}. Small definitive label clues are enough for ZSLs to achieve good accuracy. The better performance of USE with SSToT is not significant than USE with only DSM-Header (only 1\% lower), but we did not report it, because we report only the best models.

Centroid based methods are not significantly better in both Word Embedding Family (WEF) and SEF models, also indicates average word vector representation of labels do not add any value. However we find that centroid of top-k labels performs better than simple centroid in WEF, because top-k label conrains more useful semantic signals than others. Moreover, small gap in mapper methods in SEF than WEF indicates mapper models perform better in SEF than WEF.

\begin{table}[h!]
\small
    \begin{center}
    \caption{\label{tab:dsd-f1-score} DSD Task Micro-F1 scores for the best models under the ZSL families}

        \begin{tabular}{||l l c||} 
        \hline
        ZSL-Family & Model-Name & Micro-F1 \\ [0.5ex] 
        \hline\hline
        \multirow{3}{4em}{WEF} 
        & \textbf{DSE(MH+DH)-Top-1} & \textbf{0.4557}($\pm 0.0383$) \\ 
        & Glove-ATE(DH)-Top-3 &0.3785($\pm 0.0430$)   \\ 
        & ATE(DH)-Centroid-Top-9 & 0.3589($\pm 0.0231$) \\ 
        & DSE-Centroid-Top-K(MH+DH)-Top-1 & 0.3761($\pm 0.0607$) \\ 
        \hline
        \multirow{3}{4em}{SEF} 
        & \textbf{USE(SSToT)-Top-1} & \textbf{0.5142}($\pm 0.0444$) \\ 
        & USE-Mapped(MH+DH)-Top-1 & 0.4730($\pm 0.0121$) \\ 
        & USE-Mapped-Centroid(MH+DH)-Top-3 & 0.3711($\pm 0.0222$) \\ 
        & USE-Mapped-Centroid-Top-K(MH+DH)-Top-3 & 0.3711($\pm 0.0222$) \\
        \hline
        \multirow{1}{6em}{NLI} 
        & \textbf{Facebook-BART(MH)-Top-1}  & \textbf{\textit{0.5205}}($\pm 0.0196$) \\ 
        \hline
        \multirow{1}{6em}{Baseline} 
        & BERT-Fine-tuned & 0.3299($\pm 0.0246$) \\
        \hline
        \end{tabular}
    \end{center}
\end{table}

\subsubsection{Depressive Post Detection Accuracy} 
We see significant discriminatory capability of ZSL models than the baselines when their prediction scores (i.e., cosine similarity or entailment-probability) for various Depression symptoms are used to represent a Tweet and were fed to SVM 
for the task. In Table \ref{tab:dpd-f1-score} we report the best model's i.e Facebook-BART's Depression prediction capability.


\begin{table}[h!]
\small
\centering
\caption{\label{tab:dpd-f1-score}F1 scores in DPD task}
 \begin{tabular}{||l c ||} 
 \hline
 Features & F1-Score \\ [0.5ex] 
 \hline\hline
 Depression-Symptoms-Score(Facebook-BART) & \textbf{0.8113}($\pm 0.0246$) \\ 
 LIWC-Score & 0.7574($\pm 0.0304$)  \\
 Twitter-Embedding & 0.7822($\pm 0.033$)\\ 
 Random-Uniform & 0.5205($\pm 0.0257$) \\
 Majority-Class & 0.6966  \\
 [1ex] 
 \hline
 \end{tabular}
\end{table}

\subsubsection{EI-Score} 
\label{subsubsec:ei-score}

We observe that EI-score wise, sentence embedding based model (USE(SSToT)-Top-1) achieves significantly better score than all the other methods followed by word embedding based, DSE(MH+DH)-Top-1 and NLI based Facebook-BART(MH)-Top-1, See Table \ref{tab:avg-ei-score}. Interestingly, the Facebook-BART achieves significantly high accuracy for DSD task, although in-terms of explainability it's worst among the other models, which confirms the inefficacy of entailment scores compared to cosine-similarity to find out salient n-gram explanations. We also observe that ngramex and STEP EI-score usually agrees with each other, although for USE(SSToT)-Top-1, this difference is significant could be due to the fact that STEP is capable of extracting explanations which are semantically consistent compared to inconsistent ngrams often extracted by ngramex. In table \ref{tab:exp-examples}, we see two examples, where in first example STEP explanations provide high score (0.15) than the same for ngramex (0.1), the reason for EI-Score penalization for ngramex is that, the first explanation is almost the same size as the original Tweet. In the second example, ngramex EI-Score is higher (0.21) than STEP (0.18), here the EI-score penalization for STEP is because of the same reason as first example, i.e., its first explanation is similar in size to the original Tweet. However, if we see the quality of the explanations, STEP explanations are better than ngramex.

\begin{table}[h!]
\small
\centering
\caption{EI-Scores for top-3 ZSL models reported at Table \ref{tab:dsd-f1-score}}
 \label{tab:avg-ei-score}
 \begin{tabular}{||l c c||} 

 \hline
 Models & STEP EI-Score (avg.) & ngramex EI-Score (avg.) \\ [0.5ex] 
 \hline\hline
 DSE(MH+DH)-Top-1 & \textbf{0.1605}($\pm 0.0971$)  & \textbf{0.1769}($\pm 0.1125$) \\ 
 USE(SSToT)-Top-1 & \textbf{\textit{0.2439}}($\pm 0.0988$)  & \textbf{\textit{0.1978}}($\pm0.1164$)  \\
 Facebook-BART(MH)-Top-1 & 0.1261($\pm 0.1155$)  & 0.1398($\pm0.1275$) \\ [1ex] 
 \hline
 \end{tabular}
\end{table}

\begin{table}[h!]
\small
\centering
\label{tab:exp-examples}
\caption{Top 2 EI explanations for the label "Feeling Worthless" for two Tweet examples, where STEP \& ngramex disagree for top EI-Scoring ZSL model: (USE(SSToT)-Top-1)}
 \begin{tabular}{|| p{3cm}| p{2cm} | p{3.5cm} | p{3cm}||} 
 \hline
 Tweet & Condition & Exps(STEP) & Exps(ngramex) \\ [0.5ex]
 \hline\hline
 No one understands me & $EI(STEP) > EI(ngramex)$  & ``No one'', ``No one understands me'' & ``No one understands'', ``one understands me''\\
 \hline
 I feel like utter shit & $EI(STEP) < EI(ngramex)$ & ``feel like utter shit'', ``shit'' & ``I feel like'', ``feel like utter''\\
 [1ex] 
 \hline
 \end{tabular}
\end{table}

\section{Earlier Work}
Most of the earlier work in text based Depression classification can be divided into two broad categories such as, (1) Post level signs of Depression detection \cite{vioules2018detection,cheng2016psychologist} and (2) User-level signs of Depression detection \cite{yazdavar2017semi,Choudhury2013Pred}. It is to be noted that task (1) is often an important pre-requisite of task (2). Even importantly, for clinically meaningful user-level signs of Depression detection, we need to have models that can identify post level signs of clinical Depression symptoms. There have been some efforts put to date for Depression symptoms detection task, such as, \cite{mowery2017understanding,yazdavar2017semi,mowery2016towards}. However, most of these works depend on either small and labor intensive gathering of human annotated Tweets or large amount of Tweets for the same through simple rule based distant supervision mechanism which tend to gather noisy Tweets. In this work, we outline a purely ZSL approach to find the semantic similarity relationship between our samples and the label descriptors (which correspond to a certain label). Further Most of the earlier work did not consider the explainability and a need for their explainability evaluation for Depression symptoms task, which is also our primary contribution in this work.

\section{Ethical Concerns}
Our project is approved by research ethics office for all aspects of data privacy, ethics and confidentiality. To preserve anonymity, all the Tweets sample used in this paper are paraphrased  and no user identifier is provided.

\section{Conclusion}
In this paper we address a challenging task, i.e. Depression Symptoms Detection (DSD) from Text. The main challenge in this task is the scarcity of labelled data. Hence we show that using various learned representation techniques and their enhancement we can formulate a ZSL approach for this task, which performs better than a strong fine-tuned BERT-based supervised base-line for the same. Moreover, we also provide an outline of an algorithm for exploring syntax tree for sub-phrases that explains a particular Tweet for its label. Finally we propose an explainability index that can be used to evaluate the explainability capability of our Zero-Shot models and we also compare and contrast between our proposed method and a naive n-gram based approach to show their efficacy in explainability process. 

\section{Future Work}
We would like to make our EI-Score robust 
to penalize less meaningful explanations, which are abundant in n-gram based explanations. Also, we would like to employ both ZSL and STEP methods to harvest more useful data and learn robust explainable DSD models through the use of active learning.

%
%
%
\bibliographystyle{splncs04}
\bibliography{emnlp2018}
\end{document}